\documentclass{article}
\usepackage{arxivrelease}
\usepackage[margin=1in]{geometry}
\usepackage{times}
\usepackage{amsfonts}
\usepackage{amsmath}
\usepackage{booktabs} 
\usepackage{bbm}
\usepackage[draft, bookmarks=false]{hyperref}
\usepackage{url}
\usepackage{listings}
\usepackage{cleveref} 
\crefname{equation}{}{}

\usepackage{xspace}

\usepackage{color}

\usepackage{pgfplots}
\usetikzlibrary{positioning,arrows,arrows.meta,calc}
\usepgfplotslibrary{groupplots}
\tikzset{
	double arrow/.style args={#1 colored by #2 and #3}{
		-{Latex[length=2*#1,width=2*#1]},line width=#1,#2,  
		postaction={draw,-{Latex[length=2*#1-4pt,width=2*#1-4pt]},#3,line width=#1-2pt,
			shorten >=3pt}, 
	}
}

\pgfplotsset{compat=1.14}
\usepackage{floatrow}

\usepackage{verbatim}

\newcommand{\loss}{\mathcal{L}}
\newcommand{\data}{\mathcal{D}}
\newcommand{\sampleSize}{N}
\newcommand{\ensembleSize}{M}

\newcommand\Ut{U_{\text{tot}}}
\newcommand\Ua{U_{\text{ale}}}
\newcommand\Ue{U_{\text{epi}}}

\newcommand{\dirichlet}{\text{Dir}}
\newcommand{\categorical}{\text{Cat}}
\newcommand{\normal}{\mathcal{N}}
\newcommand{\uniform}{U}

\newcommand{\pBar}{\overline{p}}

\newcommand{\rmse}{\textit{RMSE}\xspace}
\newcommand{\genUncertMeas}{I}
\newcommand{\klDiv}{\text{KL}}
\newcommand{\ause}{\textit{AUSE}\xspace}
\newcommand{\se}{\textit{SE}\xspace}
\newcommand{\ece}{\textit{ECE}\xspace}
\newcommand{\acc}{\textit{acc}\xspace}
\newcommand{\conf}{\textit{conf}\xspace}

\newcommand{\softMax}{\text{soft-max}\xspace} 
\newcommand{\nll}{\textit{NLL}\xspace}
\newcommand{\ood}{\textit{OOD}\xspace}

\DeclareMathOperator{\argmax}{\text{argmax}}
\DeclareMathOperator{\E}{\mathbb{E}}

\newcommand{\toyDataMarkSize}{0.3}

\lstdefinelanguage{BibTeX}
  {keywords={%
      @article,@book,@collectedbook,@conference,@electronic,@ieeetranbstctl,%
      @inbook,@incollectedbook,@incollection,@injournal,@inproceedings,%
      @manual,@mastersthesis,@misc,@patent,@periodical,@phdthesis,@preamble,%
      @proceedings,@standard,@string,@techreport,@unpublished%
      },
   comment=[l][\itshape]{@comment},
   sensitive=false,
}

\renewcommand{\paragraph}[1]{\noindent\textbf{#1}}
\newif\ifrenderTikz
\renderTikztrue

\let\oldtikzpicture\tikzpicture
\let\oldendtikzpicture\endtikzpicture

\renewenvironment{tikzpicture}{%
    \ifrenderTikz\expandafter\oldtikzpicture%
    \else\comment%
    \fi
}{%
    \ifrenderTikz\oldendtikzpicture%
    \else\endcomment%
    \fi
}

\title{A general framework for ensemble distribution distillation}
\author{
{\bf Jakob Lindqvist
\thanks{
    \hspace{0.14cm}Equal contribution.\newline
    \texttt{\{amanda.olmin, fredrik.lindsten\}@liu.se}\newline
    \texttt{\{jakob.lindqvist, lennart.svensson\}@chalmers.se}\newline
}
} \\
Chalmers University of Technology\\
\And
{\bf Amanda Olmin \footnotemark[1]
}\\
Link\"oping University\\
\And
{\bf Fredrik Lindsten}  \\
Link\"oping University\\
\And
{\bf Lennart Svensson}   \\
Chalmers University of Technology\\
}

\begin{document}
\onecolumn{
\noindent{}This is an edited version of the paper \textit{A general framework for ensemble distribution distillation} published in \textit{2020 IEEE International workshop on machine learning for signal processing, Sept. 21–24, 2020, Espoo, Finland}.
\\
Please cite as:

\begin{lstlisting}[language=BibTeX]
@InProceedings{ens_distr_dist,
    author = {{L}indqvist, {J}. and {O}lmin, {A}. and {L}indsten {F}. and {S}vensson {L}.},
    title = {{A} general framework for ensemble distribution distillation},
    booktitle = {{MLSP}},
    year = 2020,
}
\end{lstlisting}
}
\newpage

\maketitle

\begin{abstract}
Ensembles of neural networks have shown to give better predictive performance and more reliable uncertainty estimates than individual networks.
Additionally, ensembles allow the uncertainty to be decomposed into aleatoric (data) and epistemic (model) components,
giving a more complete picture of the predictive uncertainty.
Ensemble distillation is the process of compressing an ensemble into a single model,
often resulting in a leaner model that still outperforms the individual ensemble members.
Unfortunately, standard distillation erases the natural uncertainty decomposition of the ensemble.
We present a general framework for distilling both regression and classification ensembles
in a way that preserves the decomposition.
We demonstrate the desired behaviour of our framework
and show that its predictive performance is on par with standard distillation.
\end{abstract}

%

\section{Introduction}
Recently, there has been a surge of effort in modelling and estimating the uncertainty in deep neural networks, e.g. \cite{prior_distr_distill,what_uncert,metrics:ece,WidmannLZ:2019}.
For applications ranging from autonomous vehicles to medical image-analysis, reliable uncertainty estimates are vital.
To understand the predictive uncertainty we can decompose it into model, or \textit{epistemic}, uncertainty and inherent, \textit{aleatoric}, noise in the data.
This decomposition provides a more complete picture of the uncertainty quantification and is beneficial 

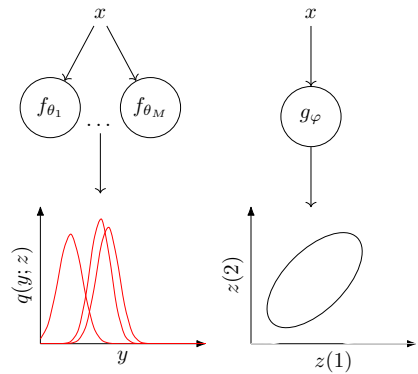
\begin{figure}[ht]
    \centering
    \pgfplotsset{compat=1.9}

\pgfmathdeclarefunction{gauss}{2}{%
  \pgfmathparse{1/(#2*sqrt(2*pi))*exp(-((x-#1)^2)/(2*#2^2))}%
}
\def\xShift{3.5cm} 
\def\xDist{0.25cm} 
\def\nodeRadius{1cm} 

\begin{tikzpicture}[scale=0.8]
    \begin{axis}[every axis plot post/.append style={
	mark=none,domain=0:25,samples=50,smooth}, 
    axis x line=bottom, 
    axis y line=left, 
    ticks=none,
    clip mode=individual,
    axis line style={-Latex[round]},
    xmin=0,
    xmax=10,
    xlabel={$y$},
    ylabel={$q(y ; z)$},
    enlargelimits=upper,
    scale=0.4] 
        \addplot[color=red] {gauss(2, 0.8)};
        \addplot[color=red] {gauss(4, 0.7)};
        \addplot[color=red] {gauss(4.5, 0.75)};
        \node[] (X) at (axis cs:4, 1.5) {$x$};
        \node[minimum size=\nodeRadius, below left =1cm and \xDist of X, draw=black, shape=circle] (f1) {$f_{\theta_1}$};
        \node[below = 1.5cm of X] (dots) {$\dots$};
        \node[minimum size=\nodeRadius, below right =1cm and \xDist of X, draw=black, shape=circle] (fM) {$f_{\theta_M}$};
        \node[below = of dots] (graphLoc) {};
        \path [->](X) edge node[left] {} (f1);
        \path [->](X) edge node[left] {} (fM);
        \path [->](dots) edge node[left] {} (graphLoc);
    \end{axis}
    \begin{axis}[every axis plot post/.append style={
	mark=none,domain=0:25,samples=50,smooth}, 
    axis x line=bottom, 
    axis y line=left, 
    ticks=none,
    clip mode=individual,
    axis line style={-Latex[round]},
    at={(\xShift, 0)},
    xmin=0,
    xlabel={$z(1)$},
    ylabel={$z(2)$}, 
    xmax=10,
    enlargelimits=upper,
    scale=0.4] 
        \addplot[color=white] {gauss(4, 0.7)};
        \node[] (X) at (axis cs:4, 1.5) {$x$};
        \node[minimum size = \nodeRadius, below = of X, draw=black, shape=circle] (dist) {$g_\varphi$};
        \node[below = of dist] (leftGraph) {};
        \draw[rotate=45] (axis cs:6, 0) ellipse (1cm and 0.5cm);
        \path [->](X) edge node[left] {} (dist);
        \path [->](dist) edge node[below] {} (leftGraph);
    \end{axis}
\end{tikzpicture}
    \caption{
    Schematic view of the general distribution distillation.
    Here, the output data is modelled with $y \vert x \sim \normal(z(1), \log(1 + e^{z(2)}))$.
    The ensemble produces several plausible predictive distributions (\emph{left}).
    The distilled model mimics this by learning a \emph{distribution} over the parameters $[z(1), z(2)]$ that captures the epistemic uncertainty in the model (\emph{right}).
    }
    \label{fig:distr_dist_schema}
\end{figure}

\noindent in applications such as active learning and reinforcement learning.

Ensembles of neural networks have shown to improve model performance and to make predictions more robust \cite{deep_ensembles} as well as to consistently provide good uncertainty estimates. 
The epistemic uncertainty is naturally characterised in an ensemble as the spread of the predictions.
Indeed, since the members are trained in an identical manner, disagreement on a given prediction means that the \emph{model} is uncertain about that prediction.

Ensemble state-of-the-art performance on \textit{out of distribution} (\ood) data,
is attributed to the ability to estimate epistemic uncertainty \cite{metrics:dataset_shift}.
However, ensembles are expensive to use at test time, both in terms of memory and computations.
It is therefore natural to consider some form of model compression that preserves the rich uncertainty description of the ensemble.

Ensemble distillation is a compression procedure where a distilled network learns to approximate the predictions of an ensemble.
The final model is often more robust and performs better than a single network trained on the same data \cite{standard_distill}.
The drawback of standard ensemble distillation (as done, e.g., by \cite{standard_distill}) is that it only considers the mean prediction of the ensemble and thereby the uncertainty decomposition is lost. 

To also capture the spread of the ensemble,
we propose to learn a \emph{distribution} over the ensemble predictions. Instead of mimicking the task of the ensemble, the training objective of the distilled network will be to predict the parameters of this distribution.
See \cref{fig:distr_dist_schema} for a schematic illustration.

Recently, a special case of this approach was proposed for classification problems,
using a Dirichlet distribution to model ensemble predictions \cite{prior_distr_distill}. 
Here we present a general framework for ensemble distribution distillation of both classification and regression models, as well as other predictive models.
Our framework is more generally applicable than previous works and allows for greater flexibility in the description of the ensemble. 
\section{Background}
\paragraph{Probabilistic predictive models}
Given a set of pairs of inputs and targets
\(
\data = \{(x_i, y_i)\}_{i=1}^\sampleSize,
\)
a probabilistic predictive model approximates the true conditional probability distribution
$p(y \vert x, \data)$,
with $q(y; f_{\theta}(x))$, where $q$ belongs to  some family of distributions parameterised by $f_{\theta}$.
In this paper, $z = f_{\theta}(x)$ is the output of a neural network that maps $x$ to a parameter vector $z$ for $q(y; z)$.

The network parameters $\theta$ are optimised in order to maximise the likelihood of data with respect to $q(y ; f_{\theta}(x))$. In practice we minimise the negative logarithm of the likelihood (\nll),
\begin{equation}
    \loss(\theta) = - \E_{p(x, y)}\left[ \log q(y ; z = f_\theta(x)) \right].
    \label{eq:ensemble_loss}
\end{equation}

\paragraph{Uncertainty quantification} 
The uncertainty in a model's prediction can be characterised using the estimated conditional probability $q(y ; f_\theta(x)) \approx p(y\vert x)$.
However, when reasoning about the uncertainty it is useful to distinguish between epistemic uncertainty in the model parameters $\theta$ and aleatoric noise in the data  \cite{what_uncert}.

For a fixed value of $\theta$, the model $q(y ; f_\theta(x))$  will only capture aleatoric uncertainty.
Conceptually, we can address this limitation with a Bayesian approach, learning a posterior distribution over the model parameters $p(\theta \vert \data)$
and expressing the predictive distribution for a data point $x^*$ as
\begin{equation}
p(y^* \vert x^*, \data) = \int \underbrace{p(y^* \vert x^*, \theta)}_{\text{aleatoric}} \underbrace{p(\theta \vert \data)}_{\text{epistemic}} d\theta.
\label{eq:uncert_decomposition_int}
\end{equation}

More specifically, we can use this approach to define the different types of uncertainty:
\begin{subequations}\label{eq:background:uncertainty-types}
\small
\begin{align}
    \text{Total: }& \Ut =
    \genUncertMeas\left[ p(y\vert x, \data) \right],
    \label{eq:total_uncert}\\
    \text{Aleatoric: }& \Ua = \E_{p(\theta \vert \data)}\!\left( \genUncertMeas \left[ p(y \vert x, \theta ) \right] \right), \label{eq:aleatoric_uncert}\\
    \text{Epistemic: }& \Ue = \Ut-\Ua, \label{eq:epistemic_uncert}
\end{align}
\end{subequations}
where $\genUncertMeas$ is some uncertainty measure, such as variance, entropy or differential entropy.

\vspace{1ex}
\paragraph{Ensembles}
Computing the posterior distribution over model parameters $p(\theta\vert\data)$ is intractable in most cases when $f_\theta$ is given by a deep neural network.
Although many approximate Bayesian methods have been proposed, e.g. \cite{Blundell15:weight-uncertainty,og_dropout}, a simple alternative is to use an ensemble of networks.
This has been found to have very competitive empirical performance \cite{metrics:dataset_shift,deep_ensembles}.

Training an ensemble with $\ensembleSize$ members means that we train $\ensembleSize$ networks independently, resulting in $\ensembleSize$ identically distributed models $\{f_{\theta_j}\}_{j=1}^{\ensembleSize}$. 
To ensure diversity in the ensemble,
random initialisation of the same network architecture and randomly sampled mini-batches are commonly considered enough.

In addition to increased performance, ensembles also provide a natural estimate of the epistemic uncertainty.
Specifically, we can use the spread of the ensemble, $\frac{1}{\ensembleSize}\sum_{j=1}^\ensembleSize \delta_{\theta_j}(\theta)$ as a plug-in replacement of the Bayesian posterior $p(\theta\mid \data)$ in \cref{eq:background:uncertainty-types}
to compute the different types of uncertainties (previously explored, e.g., by \cite{what_uncert,deep_ensembles,prior_networks,prior_distr_distill}).

\vspace{1ex}
\paragraph{Ensemble distillation}
Because of their memory usage and computational cost at test time, ensembles are good targets for model distillation \cite{Bucilua2006, standard_distill}.
In ensemble distillation, a single, distilled, model $g_{\varphi}(x)$ is trained to mimic the predictions made by the ensemble, 
after which the ensemble itself can be discarded.

Ensemble distillation is most prevalent in classification,
where the ensemble members $\{f_{\theta_j}\}_{j=1}^{\ensembleSize}$ each predict a probability vector over classes, $p_j = f_{\theta_j}$.
The distilled model $g_{\varphi}$ is also trained as a classifier using cross-entropy loss, but with the
``soft targets'' \; $\pBar = \frac{1}{\ensembleSize}\sum_j p_{j}$, rather than the hard targets $y$.
Distillation of regression models has received comparatively little attention.

\newcommand{\muDist}{\mu_{\varphi}}
\newcommand{\varDist}{\sigma^2_{\varphi}}

\section{Distribution distillation}
In this section 
we first discuss an interpretation of ``vanilla'' distillation as a \klDiv\ minimisation problem.
We then propose a general framework for distilling the distribution over the ensemble in a way that preserves the possibility of uncertainty decomposition.

\vspace{1ex}
\paragraph{Distillation as KL minimisation}
The above approach for distilling ensembles of classification models using cross-entropy loss is equivalent to interpreting the ensemble as a mixture of categorical distributions $y \sim \categorical(p)$ and to minimise the \klDiv-divergence between the distilled model and the mixture,
\begin{align}
    \klDiv \left[ 
        \categorical(y; \pBar) \|
        \categorical(y; g_{\varphi}(x)) 
    \right] = \pBar\; \log( g_{\varphi}) + C,
\label{eq:loss:cat_kl_div}
\end{align}
where $\pBar$ is the soft target from the ensemble.

Similarly, if we let both ensemble members and the distilled model parameterise some distribution over $y\vert x$ for a regression task, we can minimise the \klDiv-divergence between the mixture of the predictive distributions described by the ensemble and the distilled model. For instance, if both the ensemble members and the distilled model are assumed to be Gaussian, we get 
\begin{align}
\small
\klDiv \left[
\frac{1}{M} \sum_{j=1}^M \normal(y; z_j=f_{\theta_j}(x)) \| \normal(y; g_{\varphi}(x)) 
 \right],
\label{eq:loss:gauss_kl_div}
\end{align}
with $g_\varphi(x) = [\muDist(x), \varDist(x)]$.
For a detailed derivation, see the supplementary material.
Recent works have also used the \klDiv-divergence interpretation, \cite{Englesson2019} for classification and
\cite{hydra} for both classification and regression.
We call this approach \textit{mixture distillation}.

\vspace{1ex}
\paragraph{A general framework for distribution distillation}\label{sec:distr-distill}
The mixture distillation method captures the total uncertainty of the model but not the epistemic.
To address this limitation we propose a new framework for distillation where the distilled network predicts a distribution over the parameters $\{z_j = f_{\theta_j}(x)\}_{j=1}^M$ 
produced by the ensemble. That is, the distilled network predicts parameters for a higher-order distribution $v$ instead of 
the parameters for a distribution over the output as in mixture distillation. 
The distilled network, like the ensemble members, is trained by minimising a \nll,
but where we use the output of the ensemble $\{z_j = f_{\theta_j}(x)\}_{j=1}^{\ensembleSize}$ as the target:
\begin{align}
    \loss_{\text{DD}}(\varphi) &=
     %
     - \E_{p(x)}\!\! \left[ \frac{1}{\ensembleSize} \sum_{j = 1}^{\ensembleSize} \log v\! \left( z_j; g_{\varphi}(x) \right)\right].
    \label{eq:general_distillation_loss}
\end{align}

Note that the expectation is taken w.r.t.\ the marginal distribution over the inputs and that we use the ensemble output in place of a ground truth. Hence, the distillation process does not require annotated data. 

A key property of the proposed framework is that it is generic and applies to more than one problem class, including both classification and regression.
The generality of our framework is related to the fact that there is freedom in choosing the parameterisation of the predictive distribution $q(y;z)$.
How $z$ is interpreted can differ both between and within problem classes.
For example, in a classification setting, $z$ could represent either a \softMax-transformed probability vector or the untransformed vector in \textit{logit} space. 
This is in contrast with the work by \cite{prior_distr_distill}, that only considers distribution distillation for classification, for one choice of parameterisation.

\vspace{1ex}
\paragraph{Predictions and uncertainty quantification}
The advantage of our proposed distillation framework is that it produces a network which not only models the ensemble predictions but also its epistemic uncertainty, encoded in the distribution $v(z; g_{\varphi}(x))$.
The distilled network can 
be used to make predictions through the marginal predictive distribution,
\begin{equation}
    \tilde{q}(y; g_{\varphi}(x)) = \int q(y; z) v(z; g_{\varphi}(x)) dz.
    \label{eq:predictive_dist}
\end{equation}

Similarly to \cref{eq:background:uncertainty-types}, it can also be used for computing the total and aleatoric uncertainties,
\begin{subequations}
\begin{align}
    \text{Total: }& \Ut = 
    \genUncertMeas\left[ \tilde{q}(y; g_{\varphi}(x)) \right],
    \label{eq:dist:total_uncert}\\
    \text{Aleatoric: }& \Ua = \E_{v(z; g_{\varphi}(x))}\!\left( \genUncertMeas \left[ q(y; z) \right] \right). \label{eq:dist:aleatoric_uncert}
\end{align}
\end{subequations}
If the involved expectations are intractable we can approximate them by sampling.
Let $z_t \sim v(z; g_{\varphi}(x))$, $t = 1,\dots,T$ be independent draws from the distilled distribution.
Then
    \begin{align*}
        \Ut &\approx \genUncertMeas\left[ \frac{1}{T}\sum_{t=1}^T q(y; z_t) \right], &
        \Ua &\approx \frac{1}{T} \sum_{t=1}^T \genUncertMeas \left[ q(y; z_t) \right] .
    \end{align*}

The epistemic uncertainty is given by
\cref{eq:epistemic_uncert}.
\section{Experiments}
We evaluate our proposed framework in both regression and classification settings.\footnote[1]{Code available at github.com/jackonelli/ensemble\_distr\_distillation} 
Note that the purpose of the distillation is to compress the ensemble for efficiency. We expect that this compression comes at the price of a performance drop. Hence, the purpose of the illustration is not to show that a distilled model outperforms an ensemble, but rather that it has comparable performance at a fraction of the cost in memory and computation.

\subsection{Regression}
\begin{figure*}[ht]
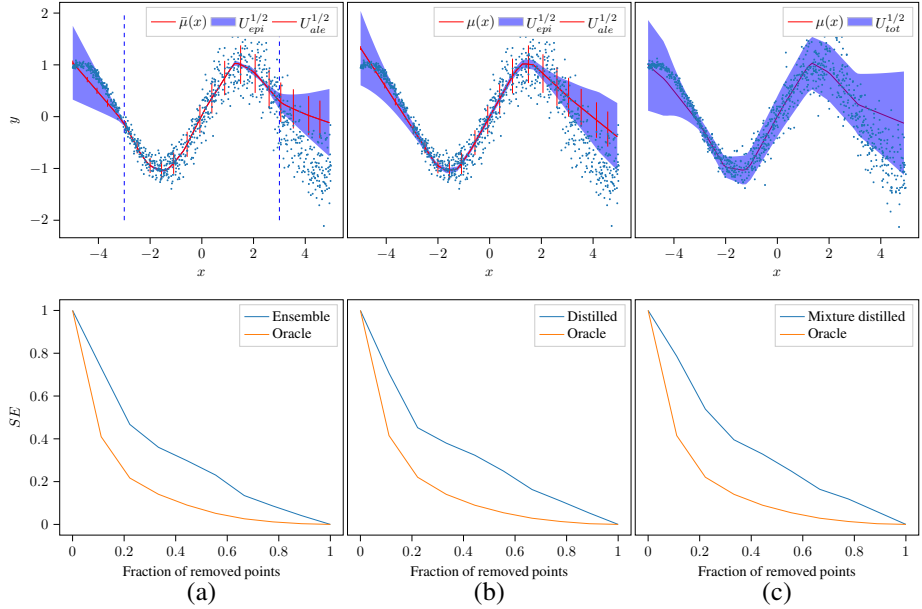

\floatbox[{\capbeside\thisfloatsetup{capbesideposition={right,top},capbesidewidth=3.8cm}}]{figure}[\FBwidth]
{\hspace{-0.3cm}\vspace{0.3cm}\caption{\textbf{Top:} Mean prediction and uncertainty estimation on toy data in \cref{eq:toy_data}.
        (a) Ensemble, trained on data on the interval $[-3, 3]$.
        (b) Our framework.
        (c) Mixture distillation.
        Distilled networks are trained only on ensemble predictions on $x$ sampled uniformly on $[-5, 5]$.
        Our framework preserves the uncertainty decomposition, whereas the mixture distillation only estimates the total uncertainty.
        \textbf{Bottom:} Sparsification plots of the toy data set for the respective models.}        \label{fig:regression:toy_example:common}}{
\begin{tikzpicture}[scale=0.55]
\definecolor{color0}{rgb}{0.12156862745098,0.466666666666667,0.705882352941177}
\definecolor{color1}{rgb}{1,0.498039215686275,0.0549019607843137}
\begin{groupplot}[
group style={group name=plot,
group size=3 by 2,
horizontal sep=0.1cm,
vertical sep=1.5cm}
]
\nextgroupplot[
legend columns=3,
legend cell align={left},
legend style={draw=white!80.0!black},
tick align=outside,
tick pos=left,
x grid style={white!69.01960784313725!black},
xlabel={\(\displaystyle x\)},
xmin=-5.49845086682181, xmax=5.46698629010061,
xtick style={color=black},
y grid style={white!69.01960784313725!black},
ylabel={\(\displaystyle y\)},
ymin=-2.33814102404719, ymax=2.20657814400225,
ytick style={color=black}
]
\input{Paper/Experiments/Regression/fig/toy_example/ensemble_mu_uncert.tikz}
\nextgroupplot[
legend columns=3,
legend cell align={left},
legend style={draw=white!80.0!black},
tick align=outside,
tick pos=left,
x grid style={white!69.01960784313725!black},
xlabel={\(\displaystyle x\)},
xmin=-5.49845086682181, xmax=5.46698629010061,
xtick style={color=black},
y grid style={white!69.01960784313725!black},
ymin=-2.33814102404719, ymax=2.20657814400225,
ytick style={color=black}
xmajorticks=false,
ymajorticks=false,
]
\input{Paper/Experiments/Regression/fig/toy_example/distilled_mu_uncert.tikz}
\nextgroupplot[
legend columns=3,
legend cell align={left},
legend style={draw=white!80.0!black},
tick align=outside,
tick pos=left,
x grid style={white!69.01960784313725!black},
xlabel={\(\displaystyle x\)},
xmin=-5.49845086682181, xmax=5.46698629010061,
xtick style={color=black},
y grid style={white!69.01960784313725!black},
ymin=-2.33814102404719, ymax=2.20657814400225,
ytick style={color=black}
xmajorticks=false,
ymajorticks=false,
]
\input{Paper/Experiments/Regression/fig/toy_example/mixture_distilled_mu_uncert.tikz}
\nextgroupplot[
legend cell align={left},
legend style={draw=white!80.0!black},
tick align=outside,
tick pos=left,
x grid style={white!69.01960784313725!black},
xlabel={Fraction of removed points},
xmin=-0.05, xmax=1.05,
xtick style={color=black},
y grid style={white!69.01960784313725!black},
ylabel={$SE$},
ymin=-0.05, ymax=1.05,
ytick style={color=black}
]
\addplot [semithick, color0]
table {%
0 1
0.111 0.731444785834719
0.222 0.467346300941728
0.333 0.360480792390979
0.444 0.297738912812318
0.555 0.230614169589372
0.666 0.134925595850052
0.777 0.0861632955193016
0.888 0.0412408818443691
1 0
};
\addlegendentry{Ensemble}
\addplot [semithick, color1]
table {%
0 1
0.111 0.409733311421693
0.222 0.216824191829413
0.333 0.140948747127818
0.444 0.090310161982688
0.555 0.0517130587292011
0.666 0.0262998626240422
0.777 0.0118536356612258
0.888 0.00340379835701472
1 0
};
\addlegendentry{Oracle}

\nextgroupplot[
legend cell align={left},
legend style={draw=white!80.0!black},
tick align=outside,
tick pos=left,
x grid style={white!69.01960784313725!black},
xlabel={Fraction of removed points},
xmin=-0.05, xmax=1.05,
xtick style={color=black},
y grid style={white!69.01960784313725!black},
ymin=-0.05, ymax=1.05,
ymajorticks=false,
]
\addplot [semithick, color0]
table {%
0 1
0.111 0.706968080047036
0.222 0.452007648552203
0.333 0.379916477200322
0.444 0.32329944392194
0.555 0.249299347888119
0.666 0.162968075364699
0.777 0.109211135406219
0.888 0.0531650202130078
1 0
};
\addlegendentry{Distilled}
\addplot [semithick, color1]
table {%
0 1
0.111 0.415010022038201
0.222 0.220495306828978
0.333 0.140568898023833
0.444 0.0896301513515062
0.555 0.0545189042944929
0.666 0.0288827800707969
0.777 0.0135183068144718
0.888 0.00326162132324048
1 0
};
\addlegendentry{Oracle}

\nextgroupplot[
legend cell align={left},
legend style={draw=white!80.0!black},
tick align=outside,
tick pos=left,
x grid style={white!69.01960784313725!black},
xlabel={Fraction of removed points},
xmin=-0.05, xmax=1.05,
xtick style={color=black},
y grid style={white!69.01960784313725!black},
ymin=-0.05, ymax=1.05,
ymajorticks=false,
]
\addplot [semithick, color0]
table {%
0 1
0.111 0.785811598334373
0.222 0.538864184201333
0.333 0.39538560199486
0.444 0.328610070974142
0.555 0.249299370776428
0.666 0.163914461951185
0.777 0.118682734430194
0.888 0.0596425058035741
1 0
};
\addlegendentry{Mixture distilled}
\addplot [semithick, color1]
table {%
0 1
0.111 0.415010022038201
0.222 0.220495306828978
0.333 0.140568898023833
0.444 0.0896301513515062
0.555 0.0545189042944929
0.666 0.0288827800707969
0.777 0.0135183068144718
0.888 0.00326162132324048
1 0
};
\addlegendentry{Oracle}
\end{groupplot}
\node[below = 0.5cm of plot c1r2.south] {(a)};
\node[below = 0.5cm of plot c2r2.south] {(b)};
\node[below = 0.5cm of plot c3r2.south] {(c)};
\end{tikzpicture}\vspace{-1.5\baselineskip}}
\end{figure*}

Regression is an under-explored topic in distillation.
Here, we demonstrate how our framework can be used in that setting.
First, we present regression distillation on a toy problem and then illustrate its performance on some real-world datasets.

\vspace{1ex}
\paragraph{Regression toy example}
The example data set is the same as used by \cite{k_gustaf} and is a sinusoidal curve with heteroscedastic noise
\begin{align}
    \label{eq:toy_data}
    y(x) = \sin(x) + \varepsilon(x), 
    \ \varepsilon \sim \normal \left(0, \frac{0.15}{1 + e^{-x}} \right).
\end{align}
An ensemble with $\ensembleSize=10$ members, each member with a single hidden layer, 
predicts $\ensembleSize$ normal distributions on the form $\normal(y; z(1), \log(1 + e^{z(2)}))$. The ensemble is trained on $N=1000$ pairs $\{(x_i, y_i)\}_{i=1}^{N}$ with $x_i$ sampled uniformly on $[-3, 3]$. To illustrate behaviour on \ood data, we evaluate the ensemble on data sampled uniformly on the larger interval $[-5, 5]$. For further details on the training, see the supplementary material.

The aleatoric and epistemic uncertainty is calculated according to \cref{eq:aleatoric_uncert,eq:epistemic_uncert},
respectively, using variance as a measure of uncertainty.
The average mean prediction and decomposed uncertainty are shown in \cref{fig:regression:toy_example:common}(a).

The ensemble is distilled to a single network, parameterising a diagonal normal distribution over $z = [z(1), z(2)]$.
This network has 2 hidden layers with 10 neurons each.
Training is done on the ensemble predictions on inputs drawn from $\uniform[-5, 5]$.
We emphasize that the distillation training is unsupervised and does not require ground truth values $y$. In addition, we note that the distribution parameterised by the distilled network differs from that of the ensemble members, since it is a distribution over parameters and not over the output of the network.
The distilled network is evaluated in the same way as the ensemble and the result is shown in \cref{fig:regression:toy_example:common}(b).
The results indicate that our framework successfully distills the ensemble while retaining its rich uncertainty description.

For comparison, we also train a network with mixture distillation.
We use the same architecture as for the distribution distillation,
but optimise the \klDiv-divergence in \cref{eq:loss:gauss_kl_div}. In 
\cref{fig:regression:toy_example:common}(c) the distilled mean and total uncertainty are shown,
but the uncertainty decomposition is no longer available.
Sparsification error (\se) measure the error decay when the most uncertain data are removed \cite{metrics:sparse_err} (see the supplementary material).
\se plots in \cref{fig:regression:toy_example:common} confirm that both distilled networks are able to capture the total uncertainty.\\
%

\vspace{1ex}
\paragraph{UCI data}
\label{sec:regression:uci}
We use the UCI data \cite{uci_data} and perform an experiment with the setup described in \cite{prob_back_prop}.
We distill an ensemble of $\ensembleSize=10$ networks.
Individual ensemble members have a single hidden layer with 50 neurons.
The distilled model has a single hidden layer of 75 neurons, trained only on ensemble predictions.

We measure root mean squared error (\rmse), \nll and area under sparsification error plot (\ause) for both models and the results are compared in \cref{tab:uci_regression}.
Each data set is split into 5 train-test folds for which both models are re-trained and tested.
The ensemble consistently outperforms the distilled model, which is expected since the objective for the distillation is to mimic the ensemble.
Still, the distilled model is performing well in all metrics,
with confidence intervals computed over independent replications largely overlapping those of the ensemble.

\begin{table*}[t]%
\begin{center}
\resizebox{1.\columnwidth}{!}{
\begin{tabular}{l | cc |cc | cc}
\toprule 
 Datasets & \multicolumn{2}{|c|}{RMSE} & \multicolumn{2}{|c|}{NLL} & \multicolumn{2}{c}{AUSE} \\
& Ensemble & Distilled & Ensemble & Distilled & Ensemble & Distilled \\ 
\midrule
 concrete    & $8.08 \pm 2.38$ & $8.64 \pm 1.82$ & $3.47 \pm 0.23$  & $3.80 \pm 0.14$  & $0.36 \pm 0.12$ & $0.34 \pm 0.08$ \\
 wine        & $0.65 \pm 0.02$ & $0.65 \pm 0.02$ & $0.99 \pm 0.01$  & $1.05 \pm 0.02$  & $0.50 \pm 0.02$ & $0.58 \pm 0.04$ \\
 yacht       & $2.86 \pm 0.26$ & $3.42 \pm 0.32$ & $3.41 \pm 0.11$  & $4.30 \pm 0.14$  & $0.28 \pm 0.02$ & $0.34 \pm 0.11$ \\
 kin8nm      & $0.11 \pm 0.02$ & $0.12 \pm 0.02$ & $-0.72 \pm 0.31$ & $-0.27 \pm 0.43$ & $0.30 \pm 0.04$ & $0.37 \pm 0.07$ \\
 power plant & $4.31 \pm 0.20$ & $4.33 \pm 0.23$ & $3.10 \pm 0.19$  & $3.67 \pm 0.41$  & $0.57 \pm 0.06$ & $0.64 \pm 0.10$ \\
\bottomrule
\end{tabular}
}
\end{center}
\caption{Results on regression benchmark datasets comparing \rmse, \nll and \ause for the ensemble and our distillation.
Lower is better for all three metrics.
}
\label{tab:uci_regression}
\end{table*}

\subsection{Classification}
The presented framework is evaluated for classification on the CIFAR-10 dataset \cite{cifar10}. We include our model in the benchmark in \cite{metrics:dataset_shift} to measure accuracy and expected calibration error (\ece) \cite{metrics:ece} on \ood data.


The distilled model predicts a diagonal normal distribution $v(z; g_\varphi(x))=\mathcal{N}(z;\mu_\varphi(x), \Sigma_\varphi(x))$ over ensemble logits $z = [z(1), \dots, z(K-1)]$, using class $K$ as a reference class. 
We base the model on a ResNet architecture  \cite{resnet} with 20 layers%
\footnote[2]{
Based on code from https://github.com/kuangliu/pytorch-cifar/blob/master/models/resnet.py} and train it using ensembles of size $M=10$ from \cite{metrics:dataset_shift}.
The \ood data used in the experiments comes from applying 16 corruptions, such as Gaussian noise and changes to the contrast, with five levels of severity to 
CIFAR-10 test images \cite{cifar10_corrupted}. The full set of corruptions are listed in the supplementary material.


%

\begin{figure*}[ht]
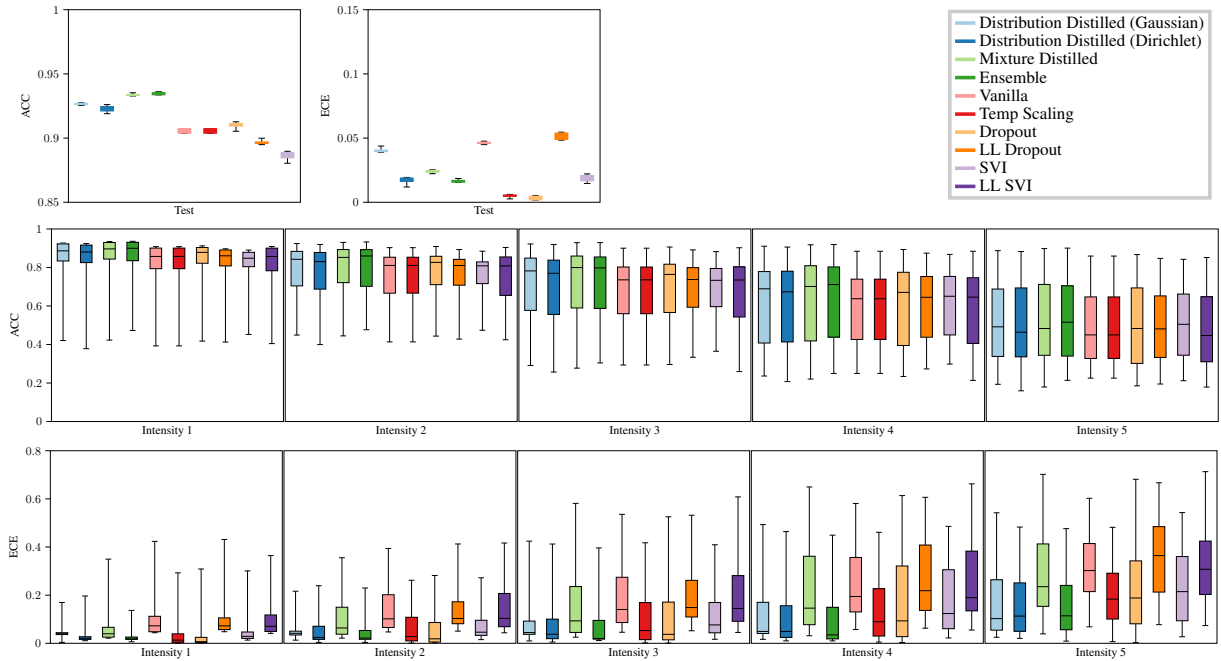

    \input{Paper/Experiments/Classification/fig/acc_ece_test.tikz}

$\!$\input{Paper/Experiments/Classification/fig/acc_benchmark_experiments_resnet20.tikz}

\input{Paper/Experiments/Classification/fig/ece_benchmark_experiments_resnet20.tikz}
     \caption{Model accuracy and \ece across CIFAR-10 test data and \ood data consisting of CIFAR-10 data distorted with 16 different corruptions applied at an intensity scale ranging from 1 to 5.
     Boxes display minimum, maximum and median together with first and third quartiles of the accuracy and \ece,
     respectively. See \cite{metrics:dataset_shift} for details on the problem setup and explanations of the competing methods (from \emph{Vanilla} to \emph{LL SVI} in the list).}
     \label{fig:classification:benchmark}
\end{figure*}

In addition to comparing the performance of our model to that of the models constructed in \cite{metrics:dataset_shift}, we train and include in the results one distribution distilled model parameterising a Dirichlet distribution according to \cite{prior_distr_distill} using a temperature annealing schedule for the \softMax function. We also include one mixture distilled model obtained with the \klDiv-divergence objective in \cref{eq:loss:cat_kl_div}. Training details can be found in the supplementary material. 
The accuracy and \ece obtained with each model over the corrupted datasets and over five repeats are displayed in  \cref{fig:classification:benchmark}. The \ece is a measure of misalignment between confidence and predictive accuracy, and is used to assess the validity of a model's uncertainty estimates (see the supplementary material).

In terms of \ece, our distribution distilled model performs comparable to the ensemble and is one of the best performing models on the corrupted data. Among the best-performing models, we also find the dropout model that bases it predictions on sampling by applying dropout during test time.
In contrast to this model, our model requires only one forward pass through the network at test time.

The distribution distilled model has a slightly lower accuracy than the ensemble and the mixture distilled model, at least on in-distribution data. 
This indicates a trade-off between cost, in terms of computation and memory as well as a trade-off between the two objectives of estimating epistemic uncertainty and predicting the mean. However, our model is more cost-efficient than the ensemble and the ability to preserve the uncertainty decomposition of the ensemble proves valuable on \ood data. The same conclusion can be drawn from the performance of the Dirichlet distilled model.

\section{Discussion and conclusion}
\label{sec:discussion}

We highlight possible extensions to the presented framework and summarise the contributions of this paper.

\vspace{1ex}
\paragraph{Possible extensions} The \nll in equation \eqref{eq:general_distillation_loss}  allows for learning 
the two tasks of making predictions and representing uncertainty,
but it lacks a way of adjusting the trade-off between the tasks. In addition, it does not offer any possibility of including the annotated data that do exist. It would be of relevance to investigate how model performance could benefit from changes to the loss function.

For classification we have parameterised $z$ in $q(y;z)$ as logits.
Using the standard parameterisation of $q$ in terms of the probability vector instead, the distillation process can have difficulties in distinguishing small (but potentially important) differences in the class probabilities. The flexibility of alleviating this problem by performing the distillation directly in logit space as opposed to specifying a temperature annealing schedule as in \cite{prior_distr_distill} is of interest for further study.


\vspace{1ex}
\paragraph{Conclusions} We have proposed a general framework for ensemble compression that maintains the rich description of predictive uncertainty,
a key advantage of ensembles.
Specifically, the compressed model estimates both epistemic and aleatoric uncertainty.
Contrary to previous work, our framework applies to both regression and classification. 
We have demonstrated that this framework can result in compressed models with performance that is highly competitive with the state-of-the-art.
Furthermore, compared to using a full ensemble, or other methods that are able to capture epistemic uncertainty
(e.g. MC dropout or VI), our distilled model is simple and efficient to use at test time and has favorable storage cost.

\subsubsection*{ACKNOWLEDGMENTS}
This work was supported by the Wallenberg AI, Autonomous Systems and Software Program (WASP) funded by the Knut and Alice Wallenberg Foundation.

\bibliographystyle{IEEEbib}
\bibliography{references}

\begin{thebibliography}{10}

\bibitem{prior_distr_distill}
A.~{Malinin}, B.~{Mlodozeniec}, and M.~{Gales},
\newblock ``{Ensemble Distribution Distillation},''
\newblock {\em arXiv:1905.00076}, Apr 2019.

\bibitem{what_uncert}
{A}. {K}endall and {Y}. {G}al,
\newblock ``{W}hat uncertainties do we need in bayesian deep learning for
  computer vision?,''
\newblock in {\em \nips}, 2017.

\bibitem{metrics:ece}
C.~{Guo}, G.~{Pleiss}, Y.~{Sun}, and K.~Q. {Weinberger},
\newblock ``{On Calibration of Modern Neural Networks},''
\newblock in {\em {\icml}}, 2017.

\bibitem{WidmannLZ:2019}
D.~Widmann, F.~Lindsten, and D.~Zachariah,
\newblock ``Calibration tests in multi-class classification: {A} unifying
  framework,''
\newblock in {\em \nips}. 2019.

\bibitem{deep_ensembles}
{B}. {{L}akshminarayanan}, {A}. {{P}ritzel}, and {C}. {{B}lundell},
\newblock ``{{S}imple and Scalable Predictive Uncertainty Estimation using Deep
  Ensembles},''
\newblock in {\em \nips}, 2017.

\bibitem{metrics:dataset_shift}
Y.~{Ovadia}, E.~{Fertig}, J.~{Ren}, Z.~{Nado}, D.~{Sculley}, S.~{Nowozin},
  J.~V. {Dillon}, B.~{Lakshminarayanan}, and J.~{Snoek},
\newblock ``{C}an you trust your model's uncertainty? {E}valuating predictive
  uncertainty under dataset shift,''
\newblock in {\em \nips}, 2019.

\bibitem{standard_distill}
{G}. {{H}inton}, {O}. {{V}inyals}, and {J}. {{D}ean},
\newblock ``{{D}istilling the Knowledge in a Neural Network},''
\newblock in {\em \nips{} Deep Learning and Representation Learning Workshop},
  2015.

\bibitem{Blundell15:weight-uncertainty}
C.~{Blundell}, J.~{Cornebise}, K.~{Kavukcuoglu}, and D.~{Wierstra},
\newblock ``{W}eight uncertainty in neural networks,''
\newblock in {\em {\icml}}, 2015.

\bibitem{og_dropout}
{Y}. {G}al and {Z}. {G}hahramani,
\newblock ``{D}ropout as a bayesian approximation: {R}epresenting model
  uncertainty in deep learning,''
\newblock in {\em {\icml}}, 2016.

\bibitem{prior_networks}
{A}. {{M}alinin} and {M}. {{G}ales},
\newblock ``{Predictive Uncertainty Estimation via Prior Networks},''
\newblock in {\em \nips}, 2018.

\bibitem{Bucilua2006}
{C}. {B}uciluǎ, {R}. {C}aruana, and {A}. {N}iculescu {M}izil,
\newblock ``{Model compression},''
\newblock in {\em Proc. 12th ACM SIGKDD Int. Conf. Knowledge discovery and data
  mining}, 2006.

\bibitem{Englesson2019}
E.~{Englesson} and H.~{Azizpour},
\newblock ``{Efficient Evaluation-Time Uncertainty Estimation by Improved
  Distillation},''
\newblock in {\em {ICML} Workshops, Workshop on Uncertainty and Robustness in
  Deep Learning}, 2019.

\bibitem{hydra}
L.~{Tran}, B.~S. {Veeling}, K.~{Roth}, J.~{Swiatkowski}, J.~V. {Dillon},
  J.~{Snoek}, S.~{Mand t}, T.~{Salimans}, S.~{Nowozin}, and R.~{Jenatton},
\newblock ``{Hydra: Preserving Ensemble Diversity for Model Distillation},''
\newblock {\em arXiv:2001.04694}, Jan 2020.

\bibitem{k_gustaf}
F.~K. {Gustafsson}, M.~{Danelljan}, and T.~B. {Sch{\"o}n},
\newblock ``{Evaluating Scalable Bayesian Deep Learning Methods for Robust
  Computer Vision},''
\newblock {\em arXiv:1906.01620}, Jun 2019.

\bibitem{metrics:sparse_err}
{C}. {K}ondermann, {R}. {M}ester, and {C}. {G}arbe,
\newblock ``{A} statistical confidence measure for optical flows,''
\newblock in {\em {\eccv}}, 2008.

\bibitem{uci_data}
D.~Dua and C.~Graff,
\newblock ``{UCI} machine learning repository,'' 2017.

\bibitem{prob_back_prop}
J.~M. {Hern{\'a}ndez-Lobato} and R.~P. {Adams},
\newblock ``{Probabilistic Backpropagation for Scalable Learning of Bayesian
  Neural Networks},''
\newblock in {\em {\icml}}, 2015.

\bibitem{cifar10}
A.~Krizhevsky,
\newblock ``Learning multiple layers of features from tiny images,'' 2009.

\bibitem{resnet}
{K}. {H}e, {X}. {Z}hang, {S}. {R}en, and {J}. {S}un,
\newblock ``{Deep Residual Learning for Image Recognition},''
\newblock in {\em {\cvpr}}, 2016.

\bibitem{cifar10_corrupted}
D.~Hendrycks and T.~Dietterich,
\newblock ``{Benchmarking neural network robustness to common corruptions and
  perturbations},''
\newblock in {\em {\iclr}}, 2019.

\bibitem{adam}
{D}.~P. {K}ingma and {J}. {B}a {L}ei,
\newblock ``{A}dam: {A} method for stochastic optimization,''
\newblock in {\em {\iclr}}, 2015.

\bibitem{metrics:original_sparse_err}
A.~Bruhn and J.~Weickert,
\newblock ``A confidence measure for variational optic flow methods,''
\newblock in {\em Geometric Properties for Incomplete Data}, pp. 283--298. Jan
  2006.

\bibitem{metrics:ause}
E.~{Ilg}, {\"O}.~{{\c{C}}i{\c{c}}ek}, S.~{Galesso}, A.~{Klein}, O.~{Makansi},
  F.~{Hutter}, and T.~{Brox},
\newblock ``{{U}ncertainty Estimates and Multi-Hypotheses Networks for Optical
  Flow},''
\newblock in {\em {\eccv}}, 2018.

\end{thebibliography}

\section{Appendix}
\subsection{Mixture distillation}
\label{appendix:direct_distillation}
An ensemble can be distilled in such a way that only the estimation of total uncertainty is preserved.

An equally weighted mixture model $p$ is constructed from the ensemble output parameters and the distilled model $q$ is optimised to produce the parameters of a single distribution,
similar to the mixture.

The similarity is measured with the \klDiv-divergence.
With the expectation in the divergence taken w.r.t. the mixture model then only one term depends on the parameters of $q$:
\begin{align}
\klDiv \left( p \| q_{\varphi} \right) &= \nonumber \\
&= \E_{p} \log p - \log q_{\varphi} \nonumber \\ 
&= - \E_{p} \log q_{\varphi} + C \\ \nonumber
\end{align}

\subsubsection{Categorical}
\newcommand{\probVec}{p}
\newcommand{\probVecBar}{\overline{p}}
For an ensemble with members proposing categorical distributions,
the distilled model predicts a categorical distribution $q_{\varphi}(y; \probVec_{\varphi}) = \probVec_{\varphi}$
that minimises the \klDiv-divergence between it and the categorical mixture.
The categorical mixture is represented by the average probability vector, the so called soft-target
$p \left( y; \{\probVec_j\}_{j=1}^M \right) = \frac{1}{M} \sum_{j=1}^M \probVec_j = \probVecBar$.
Then the \klDiv-divergence becomes

\begin{align}
\loss(\varphi) &= \klDiv \left( p \| q_{\varphi} \right) \nonumber \\
&= - \sum_{k=1}^K \probVecBar_k \log \probVec_{\varphi, k} + C \\ \nonumber
&= H(\probVecBar, \probVec_{\varphi}) + C,
\end{align}
where $H$ is the cross-entropy.

\subsubsection{Gaussian}
\renewcommand{\muDist}{\mu_{\varphi}}
\renewcommand{\varDist}{\sigma^2_{\varphi}}
For an ensemble with members proposing gaussian distributions,
the distilled model also predicts a gaussian distribution $q_{\varphi}(y; \muDist, \varDist)$ that minimises the \klDiv-divergence between it and the gaussian mixture
$p\left( y; \{\mu_j, \sigma^2_j\}_{j=1}^M \right) = \frac{1}{M} \sum_{j=1}^M \normal(y; \mu_j, \sigma^2_j)$:

{
\scriptsize
\begin{align}
\loss(\varphi) &= \klDiv \left( p \| q_{\varphi} \right) \nonumber \\
&= - \int p(y) \log q_{\varphi}(y) dy + C \\ \nonumber
&= - \frac{1}{M} \int \sum_{j=1}^M \normal(y; \mu_j, \sigma_j^2) \log \normal_{\varphi}(y; \muDist, \varDist) dy + C_1 \nonumber \\
&= - \frac{1}{M} \sum_{j=1}^M \int \normal(y; \mu_j, \sigma_j^2) \log \normal_{\varphi}(y; \muDist, \varDist) dy + C_1
\end{align}
}

The logarithm of the distilled distribution yields the following terms:
\begin{align}
&\log \normal_{\varphi}(y; \muDist, \varDist) \nonumber \\
&= \frac{(y - \muDist)^2}{\varDist} - \frac{1}{2} \log(\varDist) - \frac{1}{2} \log(2\pi) \nonumber \\
&= \frac{(y - \muDist)^2}{\varDist} - \frac{1}{2} \log(\varDist) + C_2,
\end{align}

where only the first term depends on $y$. The denominator in the first term can in turn be expanded to:
{\small
\begin{align}
(y - \muDist)^2 =
(y - \mu_j + \mu_j - \muDist)^2 \nonumber  \\
= (y - \mu_j )^2 + ( \mu_j - \muDist)^2  + 2(y - \mu_j )( \mu_j - \muDist)
\end{align}
}

With $y$ a stochastic variable distributed according to $\normal(y; \mu_j, \sigma_j^2)$,
the expectation of these terms are
\begin{align}
\E_{\normal(y; \mu_j, \sigma_j^2)} (y - \muDist)^2
= \sigma_j^2 + ( \mu_j - \muDist)^2 
\end{align}
In total the expectation of all the terms is
\begin{align}
\loss(\varphi)
&= \frac{1}{M} \sum_{j=1}^M 
\left[ \frac{\sigma_j^2 + (\mu_j - \muDist)^2 }{\varDist} \right] \nonumber \\
&+ \frac{1}{2} \log(\varDist) + C_3.
\end{align}

Again, the quadratic term can be expanded, with $\bar{\mu} = \frac{1}{M} \sum_{j=1}^M \mu_j$:
\begin{align}
(\mu_m - \muDist)^2 =
(\mu_m + \bar{\mu} - \bar{\mu} - \muDist)^2 \nonumber  \\
= (\mu_m - \bar{\mu} )^2 + ( \bar{\mu} - \muDist)^2  + 2(\mu_j - \bar{\mu})( \bar{\mu} - \muDist),
\end{align}
where the middle term is not dependent on $j$ and the last one sums to 0.

Finally:
\begin{align}
\loss(\varphi)
&= \frac{1}{\varDist M} \sum_{j=1}^M 
\left[ \sigma_j^2 + (\mu_j - \bar{\mu})^2 \right] + (\bar{\mu} - \muDist)^2\nonumber \\
&+ \frac{1}{2} \log(\varDist) + C_3.
\end{align}
\section{Training details}

\subsection{Regression training details}

In the last layer of the models, the output that predicts the variance parameter is transformed to the positive real axis with

\begin{align}
    \sigma^2 = \log(1 + \exp(z)) + c.
    \label{eq:variance_transform}
\end{align}

For the UCI data in
\cref{sec:regression:uci}, the training was sensitive to initialisation and occasionally diverged.

\subsubsection{Toy example}
Each ensemble member is trained for a 150 epochs with batch size 32. We use the \textit{Adam} optimization algorithm (\cite{adam}) with learning rate $\lambda = 0.001$.
The distilled model has two hidden layers with 10 neurons in each. We train it for 30 epochs with the same optimizer as for the ensemble

\subsubsection{UCI data}

For the ensemble training we use the same setup as in \cite{deep_ensembles}. We use the \textit{Adam} optimization algorithm (\cite{adam}) with learning rate $\lambda = 0.001$.
The distilled model has a single hidden layer with 75 neurons. We train it for 30 epochs with the same optimizer as for the ensemble.

\subsection{Classification training details}
The CIFAR-10 image data is scaled to the range [0.0, 1.0] prior to training. Out of the 50,000 training images in the training set, 40,000 are randomly sampled and used for training while the remaining 10,000 images are used for validation. For training, augmentation is used in the form of random flips (horizontal) and random crops.

The distilled model is trained for 100 epochs using the \textit{Adam} optimization algorithm \cite{adam}. The learning rate is set as $\lambda = \lambda_0 \cdot k^{-c}$, where $\lambda_0=0.001$ is the initial learning rate, $k$ is the step and with $c=0.8$. A step is taken every 20$^{\text{th}}$ epoch. 

For numerical stability during training, we make the network output $K-1$ extra positive constants (with $K=10$), $c$, and parametrise the diagonal elements of the covariance matrix according to \cref{eq:variance_transform}. During the test phase, we let $\sigma^2 = z + c$ if $z > 10$ to avoid numerical issues.


A  similar training regime as above is used for the mixture distilled and the Dirichlet distilled models but with the \softMax transformed ensemble probability vectors as targets. For the mixture distilled model, we use cross-entropy loss with the mean of the probability vectors as the target and with a \softMax temperature of $T=2.5$ during training. 

For the Dirichlet distilled model we use the training objective in \eqref{eq:general_distillation_loss} with $z_j$ ensemble probability vectors and with $v(z_j; g_\varphi(x))=\dirichlet(z_j; \alpha)$ where $\alpha$ is parameterised according to $\alpha = \exp(z/T)$. Following \cite{prior_distr_distill}, we use a temperature annealing schedule starting at a \softMax temperature of $T=10$. The temperature is held constant for the first 50 epochs and is thereafter decreased by a factor $\tau = 0.95$ every epoch until $T=1$. In addition to the temperature annealing schedule, we apply \textit{central smoothing} to the ensemble member output as done in \cite{prior_distr_distill}

\begin{align}
    z_j = (1 - \gamma)z_j + \frac{\gamma}{K}
\end{align}

with $\gamma = 10^{-4}$. 

\if 
For the distilled models trained on OOD data, we add 5,000 randomly sampled data points from four of the corrupted data sets, using only intensity 2 corruptions. The following corruptions were (randomly) chosen: contrast, frost, Gaussian blur and impulse noise. The additional training data was added after 60 epochs and the models were trained for 150 epochs in total. We employed the same learning rate schedule as previously introduced.
\fi

\subsubsection{OOD data}
The corrupted CIFAR-10 data \cite{cifar10_corrupted} used for the CIFAR-10 out-of distribution experiments includes the following 16 corruptions

\begin{itemize}
  \setlength\itemsep{0.01em}
    \item Brightness
    \item Contrast
    \item Defocus blur
    \item Elastic transform (stretch/contract regions of image)
    \item Fog
    \item Frost
    \item Gaussian blur
    \item Gaussian noise
    \item Glass blur
    \item Impulse noise ("salt-and-pepper" noise, colour analogue)
    \item Pixelate
    \item Saturate
    \item Shot noise (Poisson noise)
    \item Spatter
    \item Speckle noise
    \item Zoom blur
\end{itemize}

The corruptions are applied to the CIFAR-10 test data set of 10,000 data points on a severity scale ranging from 1 to 5.

\subsection{Metrics}

\subsubsection{Sparsification plots and \ause}
\label{sec:meas:ause}
\newcommand{\startTestInd}{1}
\newcommand{\stopTestInd}{N}
Sparsification plots \cite{metrics:original_sparse_err,metrics:sparse_err} visualise the quality of the total uncertainty estimated by a regression model
that estimates both a regression estimate $\hat{y}(x)$ and a total uncertainty $I[p(y\vert x)]$.

The regression estimates are ordered from most to least estimated uncertainty,
where uncertain estimates are expected to have a larger error.
The average error is calculated for a sequence of subsets, where each new subset removes a larger fraction of the most uncertain estimates.
Ideally, larger uncertainties should correspond to larger errors (on average) and removing points with the most uncertain predictions should therefore reduce the average error.

To get a comparable score, errors are normalised to one and measured relative to an \textit{oracle},
which orders the estimates by the actual error.
The difference between the oracle and model sparsification is called sparsification error (\textit{SE}).
The area under the \se (\ause), is a single value measuring the quality of the uncertainty estimates
\cite{metrics:ause}.

\subsubsection{Expected Calibration Error}
\label{sec:uncert_meas:ece}
\ece evaluates how well the average confidences of the predictive model matches the corresponding accuracy,
reflecting how well-calibrated the model is \cite{metrics:ece}.
Given a model $\tilde{q}$, the \ece is calculated over buckets $B_{s} = \{i \in [1, N]: \ \tilde{q}(\hat{y}_i \vert x_i) \in (\rho_{s}, \rho_{s+1}] \}$ of the set of observations $\{x_{i}\}_{i=1}^{N}$ as
\begin{align}
    ECE = \sum_{s=1}^{S}\frac{|B_{s}|}{N}|acc(B_{s})-conf(B_{s})|,
\end{align}
with,
\begin{align}
    \acc(B_{s}) = \frac{1}{|B_{s}|}\sum_{i \in B_{s}}\mathbbm{1}(y_{i}=\hat{y}_{i}), \nonumber \\
    \conf(B_{s}) = \frac{1}{|B_{s}|}\sum_{i \in B_{s}} \tilde{q}(\hat{y}_i \vert x_i), \nonumber
\end{align}
where $\hat{y}_{i} = \argmax_{y} \ \tilde{q}(y \vert x_{i})$ are model predictions and $\mathbbm{1}(\cdot)$ is an identity function. We let $\mathbb{\rho}=\{\rho_{s}\}_{s=1}^{S}$ be quartiles plus minimum and maximum (0 and 1). 

\end{document}